\documentclass[10pt,twocolumn,letterpaper]{article}

\usepackage{cvpr}              

\usepackage{graphicx}
\usepackage{amsmath}
\usepackage{amssymb}
\usepackage{booktabs}
\usepackage{bm}
\usepackage{mathrsfs}
\usepackage{multirow}
\usepackage[accsupp]{axessibility}
\usepackage[pagebackref,breaklinks,colorlinks]{hyperref}

\usepackage[capitalize]{cleveref}
\crefname{section}{Sec.}{Secs.}
\Crefname{section}{Section}{Sections}
\Crefname{table}{Table}{Tables}
\crefname{table}{Tab.}{Tabs.}

\def\cvprPaperID{8688} 
\def\confName{CVPR}
\def\confYear{2022}

\begin{document}

\title{Time3D: End-to-End Joint Monocular 3D Object Detection and Tracking for Autonomous Driving}

\author{Peixuan Li\\
PP-CEM \& Rising Auto\\
{\tt\small lipeixuan@saicmotor.com}
\and
Jieyu Jin\\
PP-CEM \& Rising Auto\\
{\tt\small jinjieyu@saicmotor.com}
}
\maketitle

\begin{abstract}
   While separately leveraging monocular 3D object detection and 2D multi-object tracking can be straightforwardly applied to sequence images in a frame-by-frame fashion, stand-alone tracker cuts off the transmission of the uncertainty from the 3D detector to tracking while cannot pass tracking error differentials back to the 3D detector. In this work, we propose jointly training 3D detection and 3D tracking from only monocular videos in an end-to-end manner. The key component is a novel spatial-temporal information flow module that aggregates geometric and appearance features to predict robust similarity scores across all objects in current and past frames. Specifically, we leverage the attention mechanism of the transformer, in which self-attention aggregates the spatial information in a specific frame, and cross-attention exploits relation and affinities of all objects in the temporal domain of sequence frames. The affinities are then supervised to estimate the trajectory and guide the flow of information between corresponding 3D objects. In addition, we propose a temporal
   -consistency loss that explicitly involves 3D target motion modeling into the learning, making the 3D trajectory smooth in the world coordinate system. Time3D achieves 21.4\% AMOTA, 13.6\% AMOTP on the nuScenes 3D tracking benchmark, surpassing all published competitors, and running at 38 FPS, while Time3D achieves 31.2\% mAP, 39.4\% NDS on the nuScenes 3D detection benchmark.
\end{abstract}

\section{Introduction}
\label{sec:intro}
\begin{figure}[htb]
\setlength{\abovecaptionskip}{0pt}
	\begin{center}
		\includegraphics[width=1\linewidth]{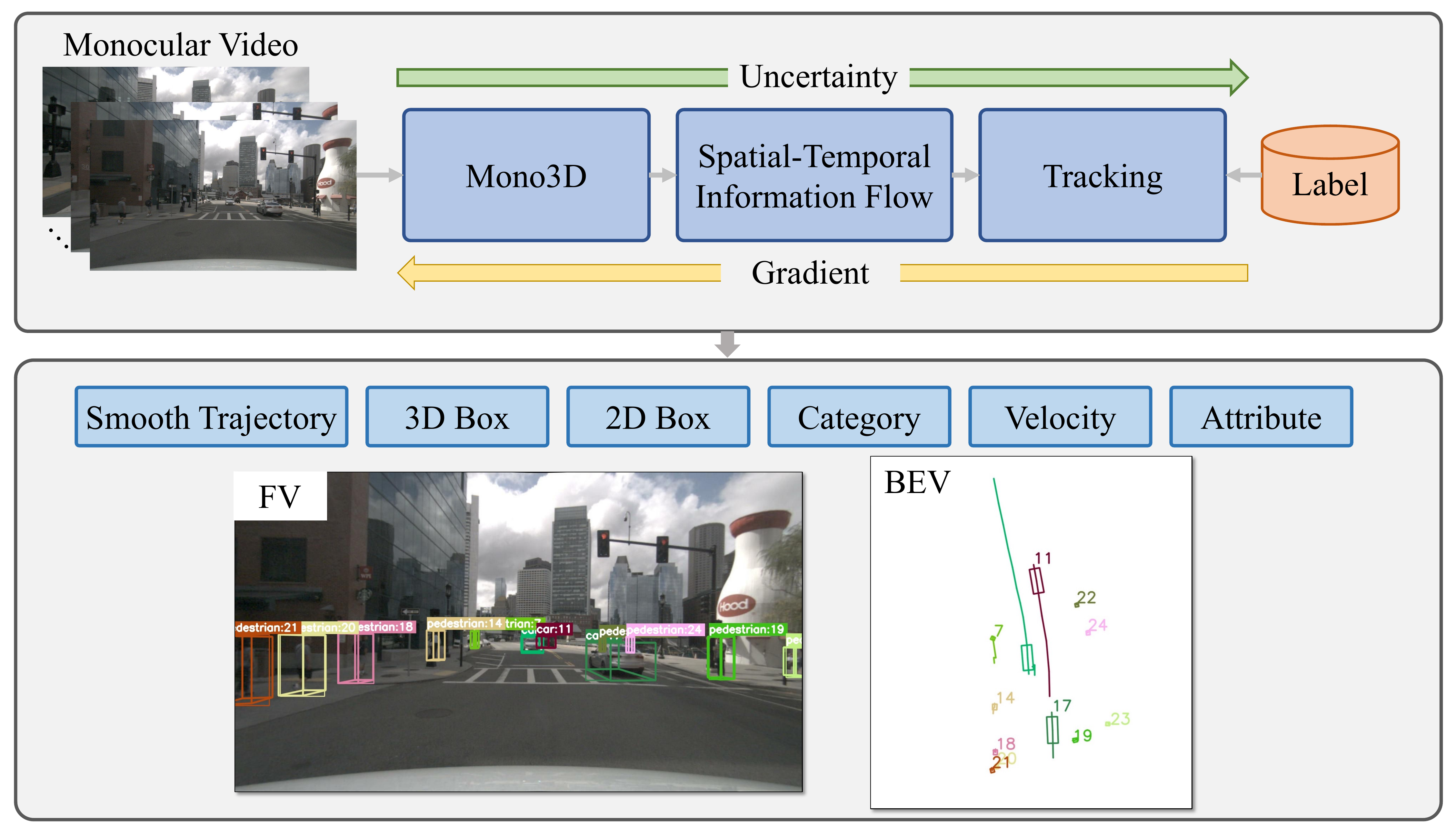}
	\end{center}
	\caption{\textbf{Illustration of the proposed Time3D.} Given a monocular video sequence, Time3D joint learn monocular 3D object detection and 3D tracking, and output smooth trajectories, 2D boxes, 3D boxes, category, velocity, and motion attribute. Time3D is trained end-to-end, so it can feed forward the uncertainty and backward the error gradient.}
	\label{fig:abstract}

\end{figure}
3D object detection is an essential task for Autonomous Driving. Compared with LiDAR system, monocular cameras are cheap, stable, and flexible, favored by mass-produced cars~\cite{li2020rtm3d, li2020monocular,brazil2019m3drpn,conf/cvpr/DingHYWSLL20}. However, monocular 3D object detection is a natural ill-posed problem for the lack of depth information, making it difficult to estimate an accurate and stable state of the 3D target~\cite{li2020rtm3d, mousavian20173d}. A typical solution is to smooth the previous and current state through 2D multiple object trackers(MOT)~\cite{DBLP:conf/iccel/LiuSC18,DBLP:conf/eccv/BrazilPLS20}. 

Following the “tracking-by-detection” paradigm~\cite{DBLP:conf/icip/BewleyGORU16,DBLP:conf/cvpr/TangAAS17,DBLP:conf/iccv/Xu0ZH19}, a widely used contemporary strategy first computes past trajectory representations and detected objects. Later, that association module is used to compute the similarity between the current objects across the past frames to estimate their tracks. Most of the existing methods following this pipeline try to build several different cues, including re-identification(Re-ID) models~\cite{DBLP:journals/pami/RahmaniMS18,DBLP:journals/tnn/ZhangLYZTL17,DBLP:conf/bmei/ZhangZDZY18}, motion modeling~\cite{DBLP:conf/icassp/FuFNC17,DBLP:conf/avss/KutschbachBES17}, and hybrid models~\cite{DBLP:journals/pami/BaeY18}. Currently, most of these models are still hand-crafted so that the corresponding tracker can only follow the detector independently. The recent studies of 2D MOT try to establish a deep learning association in tracking~\cite{DBLP:journals/pami/SunASMS21,DBLP:conf/cvpr/BrasoL20,DBLP:conf/eccv/ZhouKK20}. However, these methods still encounter three shortcomings in autonomous driving scenarios: \textbf{1)} They treat detection and association separately, in which stand-alone tracking module cuts off the transmission of the uncertainty from the 3D detector to tracking while cannot differential pass error back to the 3D detector. \textbf{2)} The objects from the same category often have similar appearance information and often undergo frequent occlusions and different speed variations in the autonomous driving scenario. They failed to integrate these heterogeneous cues in a unified network. \textbf{3)} They estimate the trajectory without directly constraining the flow of appearance and geometric information in the network, which is crucial to the trajectory smoothness, velocity estimation, and motion attribute(e.g., parking, moving, or stopped for cars).

We propose to combine 3D monocular object detection and 3D MOT into a unified architecture with end-to-end training manner, which can: \textbf{(1)} predict 2D box, 3D box, Re-ID feature from only monocular images without any extra synthetic data, CAD model, instance mask, or depth map. \textbf{(2)} encode the compatible feature representation for these cues. \textbf{(3)} learn a differential association to generate trajectory by simultaneously combining heterogeneous cues across time. \textbf{(4)} guide the information flowing across all objects to generate a target state with temporal consistency. To do so, we first modify the anchor-free monocular 3D detector KM3D~\cite{li2020monocular} to learn the 3D detector and Re-ID embedding jointly, following the "joint detection and tracking" paradigm~\cite{DBLP:journals/ijcv/ZhangWWZL21,DBLP:conf/eccv/WangZLLW20}, so that it can generate 2D box, 3D box, object category, and Re-ID features, simultaneously. To design the compatible feature representation for different cues, we propose to transform the parameters of the different magnitude of the 2D box and 3D box to the unified representation, 2D corners, and 3D corners, in which the geometric information can be extracted as a high-dimensional feature from corner raw coordinates by the widely used PointNet~\cite{qi2016pointnet} structure. Fig.~\ref{fig:abstract} illustrate the pipeline of Time3D.
 
Reviewing MOT, we found that data association is very similar to the Query-Key mechanism, where one object is a query, and another object in a different frame is the key. Its feature in different frames is highly similar for the same object, enabling the query-key mechanism to output a high response. We, therefore, propose a Transformer architecture, a widely-used entity of query-key mechanism. Inspired by RelationNet~\cite{journals/corr/abs-1711-11575}, the self-attention aggregates features from all elements in a frame to exploit spatial topology, which is automatically learned without any explicit supervision. The cross-attention computes the target affinities across different frames, and its query-key weights are supervised to learn tracks via a unimodal loss function. The finally temporal-spatial features output the velocity, attribute, and 3D box smoothness refinement. In addition, we propose temporal-consistency loss that constrains the temporal topology of objects in the 3D world coordinate system to make the trajectory smoother.

To summarize, the main contributions in this work are the following: \textbf{(1).} We propose a unified framework to jointly learn 3D object detection and 3D multi-object tracking by combining heterogeneous cues in an end-to-end manner. \textbf{(2).} We propose an embedding extractor to make geometric and appearance information compatible by transforming the 2D and 3D boxes to unified representations.
\textbf{(3).} We propose a temporal-consistency loss to make the trajectory smoother by constraining the temporal topology.
\textbf{(4).} Experiments in the nuScenes 3D tracking benchmark demonstrate that the proposed method achieves the best tracking accuracy comparing other competitors by large margins while running in real-time(26FPS).
\section{Related Work}
\textbf{Monocular 3D Object Detection}
\emph{Monocular 3D object detection} is a naturally ill-posed problem that is more complicated than 2D detection. The central problem is the lack of depth from only the perspective images. To address this challenge, the variants of Pseudo-LiDAR~\cite{wang2019pseudo} first estimate a depth map and then transform the depth map into point clouds following a LiDAR-based methods~\cite{2018SECOND,2017VoxelNet,yang2018pixor} to detect 3D objects. Another method~\cite{conf/eccv/MaLXZZO20} remove the transforming step and directly use depth to guide estimation in CNN. Inspired by the 2D anchor-based detection~\cite{brazil2019m3drpn,DBLP:conf/eccv/BrazilPLS20}, put the 3D anchors in 3D space, and filter anchors with the 2D box helping. In order to avoid direct estimation of the depth value, the variants of RTM3D~\cite{li2020rtm3d,li2020monocular} tried to combine the CNN and geometry projection to infer depth or position. Although these methods perform well on a single image, autonomous driving scenarios are often sequential recognition tasks, and processing objects independently may cause sub-optimal results. 
For this reason, \cite{DBLP:conf/eccv/BrazilPLS20,DBLP:conf/iccv/HuCWLSKDY19} introduce MOT module to monocular 3D object detection to uses temporal information to predict stable 3D information. \cite{DBLP:conf/eccv/BrazilPLS20} takes kinematic motion into the Kalman Filter as the main tracking module, which independently runs the detection and tracking model, and only the hand-crafted geometric features are used for data association. \cite{DBLP:conf/iccv/HuCWLSKDY19} include more geometric information and appearance for data association, however, it leverages a hand-crafted data association strategy and not contain a spatial-temporal information flow for box consistency constrain, which is more important to downstream tasks of autonomous driving.  

\textbf{Multi-Object Tracking}
\emph{Multi-object tracking} has been explored extensively in the 2D scope, in which most trackers follow the tracking-by-detection paradigm~\cite{DBLP:conf/iccel/LiuSC18, DBLP:conf/wacv/FangXLS18,DBLP:conf/cvpr/Leal-TaixeCS16}. These models first predict all objects by a powerful 2D detector~\cite{ren2015faster,redmon2016you,zhou2019objects} in each frame, and then link them up by data association strategy.  \cite{DBLP:conf/iccel/LiuSC18} associate each box through the Kalman filter with the distance metric of box IoU.  \cite{DBLP:conf/icip/WojkeBP17} add the appearance information from a deep network to~\cite{DBLP:conf/iccel/LiuSC18} for a more robust association, especially for occlusion objects. More recent approaches~\cite{DBLP:conf/icra/SharmaAMK18,DBLP:conf/cvpr/TangAAS17,DBLP:conf/iccv/Xu0ZH19} follow these two methods and focus on to increase the robustness of data association. These approaches separate tracking into detection and data association stages, preventing the uncertainty flow and end-to-end training. 
A recent trend in MOT is to reformulate existing trackers into the combination of both tasks in the same framework. Sun et al. ~\cite{DBLP:journals/pami/SunASMS21} use a siamese network with the paired frame as input and predict the similarity score of detections. Guillem~\cite{DBLP:conf/cvpr/BrasoL20} proposes a graph partitioning method to treat the association problem as an edge classification problem. The disadvantages of these models are that they treat detection and association separately, learn limited cues, have a complicated structure, and are not practical in autonomous driving scenes. We jointly learn 3D detection and 3D tracking by exploiting heterogeneous cues while running in real-time.

\section{Proposed Approach}
Given a monocular image $I_t$ at time $t$ 
an autonomous vehicle needs to perceive the location $\bm{P}_t = \{\bm{P}_t^i \in \mathbb{R}^3|i=1,\dots,n_t\}$, dimension $\bm{D}_t = \{\bm{D}_t^i \in \mathbb{R}^3|i=1,\dots,n_t\}$, orientation $\bm{R}_t = \{\bm{R}_t^i \in \mathbb{R} |i=1,\dots,n_t\}$, velocity $\bm{V}_t = \{\bm{V}_t^i \in \mathbb{R}^3 |i=1,\dots,n_t\}$, attribute$\bm{A}_t = \{\bm{A}_t^i \in \mathbb{R}^3 |i=1,\dots,n_t\}$, and smooth trajectory $\bm{\mathcal{T}}_t = \{\bm{\mathcal{T}}_t^i|i=1,\dots,n_t\}$ of $n_t$ objects around the scene to make motion planning and control. 

\begin{figure}[htb]
\setlength{\abovecaptionskip}{0pt}
	\begin{center}
		\includegraphics[width=0.9\linewidth]{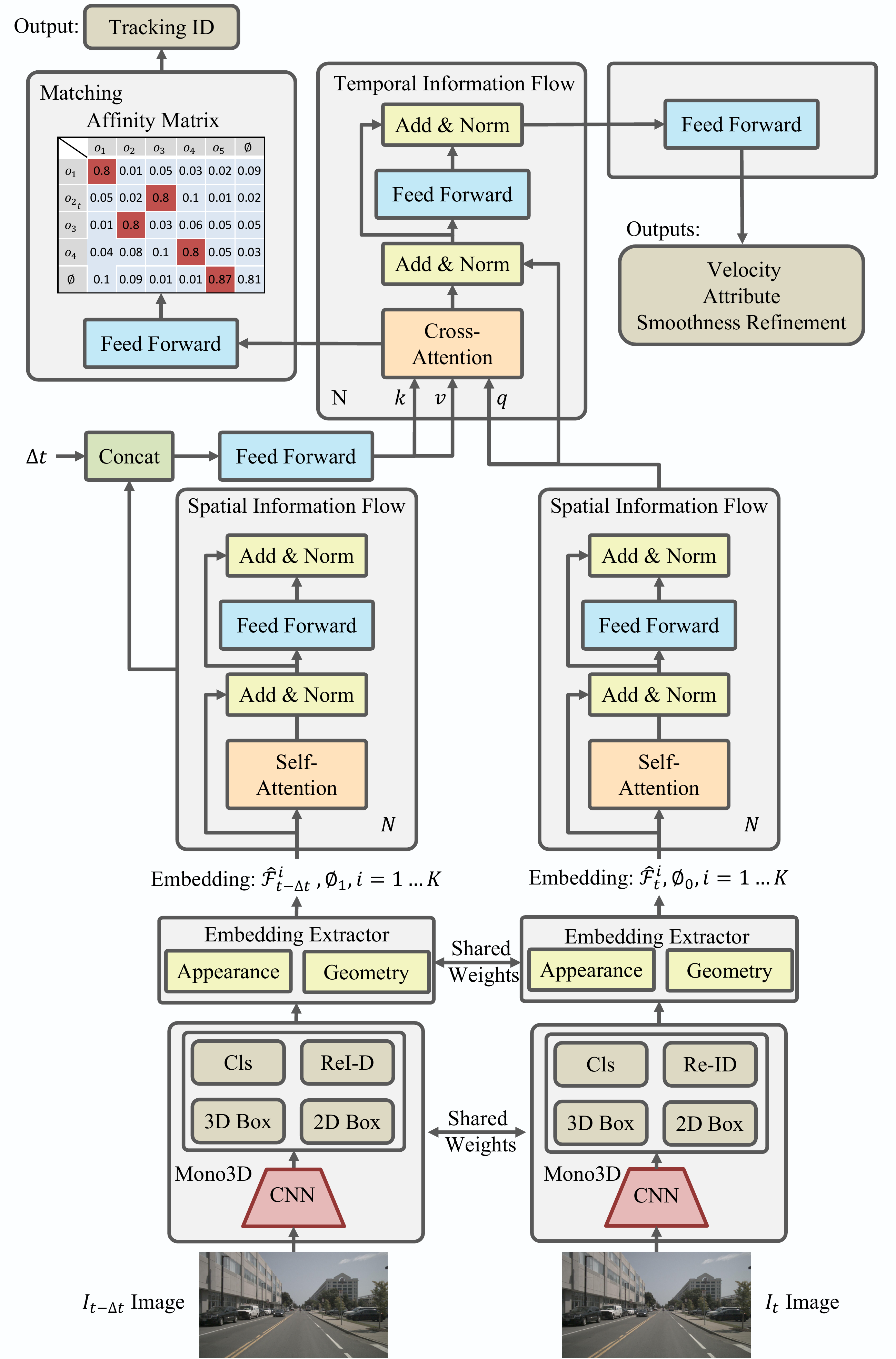}
	\end{center}
	\caption{\textbf{The architecture details of Time3D.} First,
the current and previous frame images are input to Mono3D to estimate Top $K$ objects with category, 2D Box, 3D Box, and Re-ID feature. Then, the current and previous cues are fed into the extractor of heterogeneous cues embedding to generate appearance and geometric embedding. Next, learning objects embedding are propagated to each other in a spatial domain through spatial information flow. Finally, the temporal information flow matches the same object across frames to compute affinity matrix for estimation of tracks, while output velocity, motion attribute, and box smoothness refinement.}
	\label{fig:framework}
\end{figure}

Fig.~\ref{fig:framework} illustrate the whole architecture details of Time3D. Tim3D only takes monocular video images as input and consists of the following steps: \textbf{1).} An fast and accuracy \textbf{Monocular 3D Object Detector} in JDE mode~\cite{DBLP:conf/eccv/WangZLLW20} is designed to obtain 2D boxes, 3D boxes, category, and Re-ID embedding for every frame. \textbf{2).} An \textbf{Heterogeneous Cues Embedding} module that encodes appearance and geometric features to compatible feature representation.  \textbf{3).} A \textbf{Spatial-Temporal Information Flow} module that propagates the information of all objects across frames to each other estimates the similarity to generate 3D trajectory and aggregates the geometry relative relationship in the world coordinate system to estimate velocity, attribute, and box smoothness refinement.
\subsection{Monocular 3D Object Detection}

Our monocular 3D object detector takes an image $\bm{I}_t$ at certain time $t$ as input, and outputs objects 2D box $\bm{b}^i_t = \{ x_t^i, y_t^i, w_t^i, h_t^i \}, i=1, \dots, n_t$,
3D box $\boldmath{\mathcal{B}}^i_t = \{ X_t^i, Y_t^i, Z_t^i, W_t^i, H_t^i, L_t^i \}, i=1, \dots, n_t$, category $\bm{Cls}^i_t = \{ Car, Pedestrian, \dots \}$, and ReID embedding $\bm{ReID}_t^i$.
Specifically, we employ KM3D~\cite{li2020monocular} as our monocular 3D detector, which predicts dimension, orientation, and nine perspective corners following a differentiable geometric reasoning module(GRM) for position estimation. KM3D is a camera-independent method that can handle a variety of
cameras with different intrinsic and viewing characteristics, making it more practical. Following the FairMOT~\cite{DBLP:journals/ijcv/ZhangWWZL21}, we add a Re-ID head paralleled to other detection heads, which focuses on generating distinguish features of different objects. We implement the same convolution layer with other heads but output a 256 dimension vector to extract Re-ID features in each object's 2D center.

\subsection{Heterogeneous Cues Embedding}\label{sec:embedding}
%
An ideally data association is expected to extract embedding of multiple cues(\emph{e.g.}. appearance and geometric) over a long period. However, the appearance features (\emph{e.g.} Re-ID feature) are in vector space, the geometric features (\emph{e.g.} location, dimension, and orientation) are in Euclidean space, making them difficult to be combined in a unified network. As a result, only one of them was used in previous approaches~\cite{DBLP:journals/pami/RahmaniMS18,DBLP:journals/tnn/ZhangLYZTL17,DBLP:conf/bmei/ZhangZDZY18,DBLP:conf/icassp/FuFNC17,DBLP:conf/avss/KutschbachBES17}, leading
to sub-optimal results. In this paper, we elegantly encode the compatible representations of appearance,  geometric, and motion information.  

Specifically, for every 2D box $\bm{b}_t^i$ and 3D box $\bm{B}_t^i$ in image $\bm{I}_t$, we first transform their parameters to the 2D corners $\mathcal{C}_2(\bm{b}_t^i) \in \mathbb{R}^{4 \times 2}$ and 3D corners $\mathcal{C}_3(\bm{B}_t^i) \in \mathbb{R}^{8 \times 3}$. 
These corners are flattened and then fed into light-weight PointNet~\cite{qi2016pointnet} structure, that compose of only 3 layers MLP and \emph{MaxPooling} to generate geometric $\bm{G}_t^i \in \mathbb{R}^{d}$ features with $d$ feature dimension. 

In addition to Re-ID in the appearance features, we also add category cues, which can be further used to constrain the similarity of the same object between different frames. We present the category information from an MLP layer applied on the one-hot feature layer of the cls head in our monocular 3D detector. Then we simple add the category feature and Re-ID  $\bm{ReID}_t^i$ to generate appearance feature $\bm{A}_t^i$. 

Motion modeling is time-aware processing that depends on the relative time between frames. We put motion modeling behind the spatial information flow in order to reduce calculation. We detail it in Sec.~\ref{sec:spif} and Sec.~\ref{sec:it}.
\subsection{Spatial-Temporal Information Flow}\label{sec:spif}

This section reviews the transformer structure first and then introduces how to apply it to spatial-temporal information flow.

Transformer~\cite{2017Attention} was first introduced as a new network based on attention mechanisms for machine translation and then was formulated to process computer vision tasks by ViT~\cite{DBLP:conf/iclr/DosovitskiyB0WZ21} and DETR~\cite{conf/eccv/CarionMSUKZ20}. 
For a query $\bm{\mathcal{Q}}$, key $\bm{\mathcal{K}}$, and value $\bm{\mathcal{V}}$, we simple present the multi-head attention as:
    \begin{equation}
	\begin{aligned}
	\bm{Y} = MultiHeadAttn(\bm{\mathcal{Q}}, \bm{\mathcal{K}}, \bm{\mathcal{V}}, \bm{PE})
	\end{aligned}
    \end{equation}
 where $\bm{PE}$ is the positional encoding function to eliminate the permutation-invariant. 
 This attention is called self-attention if the input query and key are the same, otherwise cross-attention. The transformer stacks self-attention, normalization, and feed-forward layers in the encoder and decoder, and the cross-attention focuses on the interaction between them. We refer the reader to the literature~\cite{2017Attention} for more detailed descriptions.  

The transformer architecture can be naturally extended to the spatial-temporal information, in which the self-attention propagate the objects' information at a certain time, and cross-attention aggregate objects information across times. The structure of the spatial information flow is shown at the bottom of Fig.~\ref{fig:framework}. We first extract top-$K$ center points in image $\bm{I}_t$ from main center head of our 3D detector, and index its corresponding appearance features $\bm{A}_{t}^i$, and geometric $\bm{G}_t^i$ (details in Sec. \ref{sec:embedding}) following the concatenation with a MLP layer to generate input embedding $\bm{\mathcal{F}}_{t}^i$.
$K$ is set to be larger than the typical number $n_t$ of objects in an image( \emph{e.g.} 128 in nuScenes dataset). 
So that the spatial information flow can be summarized as:
    \begin{equation}
	\label{quta}
	\begin{aligned}
	\bm{\Hat{\mathcal{F}}}_{t}^i = MultiHeadAttn(\bm{\mathcal{F}}_{t}^i, \bm{\mathcal{F}}_{t}, \bm{\mathcal{F}}_{t}, \bm{0})
	\end{aligned}
    \end{equation}
We set the positional encoding as $\bm{0}$ for the reason that the geometric feature already contains the position information. The exact structure with different weights can be applied in another frame $\bm{I}_{t- \Delta t}$ to generate its propagation feature $\bm{\hat{\mathcal{F}}}_{t- \Delta t}$.

The spatial information flow module process strictly follows self-attention to propagation information and encodes the spatial topology, which has been proved to be adequate to improve object detection by RelationNet~\cite{journals/corr/abs-1711-11575}.

\begin{figure*}[htb]
\setlength{\abovecaptionskip}{0pt}
	\begin{center}
		\includegraphics[width=0.9\linewidth]{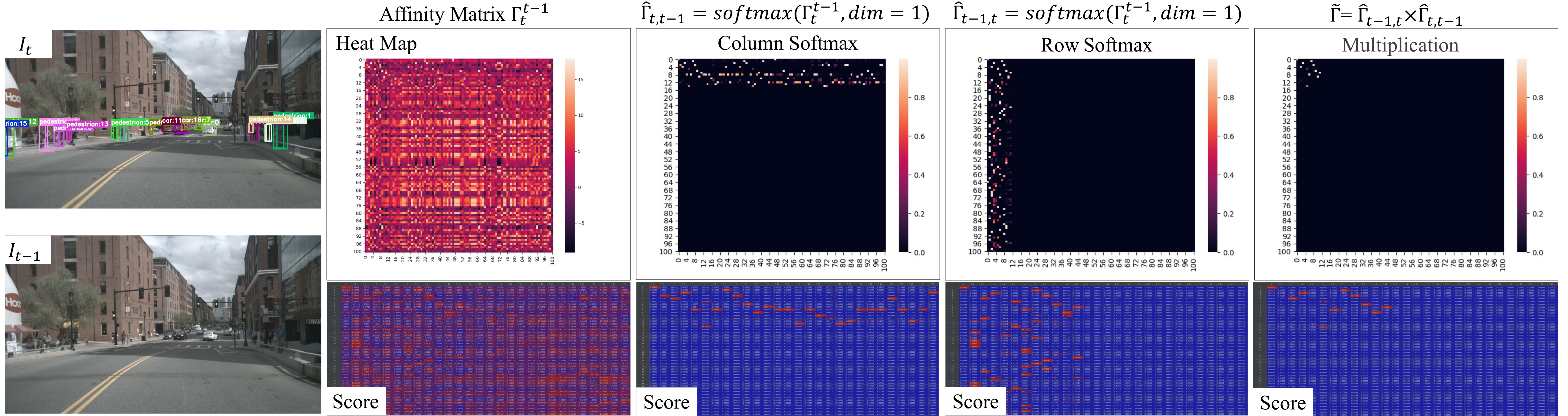}
	\end{center}
	\caption{\textbf{Learnt unimodal affinity matrix.} Our simple tracking loss can learn a unimodal affinity matrix, which can perform a one-to-one matching.}
	\label{fig:trackingscore}
	\vspace{-0.4cm}
\end{figure*}

The structure of the temporal flow module is shown at the top of Fig.~\ref{fig:framework}.
The temporal flow module aggregates the information from paired frame $\bm{I}_{t- \Delta t}$ and $\bm{I}_{t}$ by using multi-head cross-attention in the form of residual. In cross-attention, the dot-product weight explores the relationship of paired detection objects in different frames. It represents the consistency in probabilistic from 0-1 normalized by \emph{softmax}, where 0 is defined as different objects, and 1 is the same object. This probabilistic naturally contains matching information and can be used as a similarity score for tracking directly. In order to prevent the ID switch, we design Time3D as a semi-global association, so we also need to capture time information. We simple \emph{concat} the relative time $\Delta t$ and $\bm{\hat{\mathcal{F}}}_{t- \Delta t}$ following a FFN layer to generate time-aware propagation feature $\bm{\Check{\mathcal{F}}}_{t- \Delta t}$, that also called motion modeling. We formulate the generation of affinity matrix as:
    \begin{equation}
	\label{quta}
	\begin{aligned}
    \bm{\Gamma}(i, j)_t^{t-\Delta t} = \bm{FFN} \left( \bm{W}_q^i\bm{\Hat{\mathcal{F}}}_{t}^i*\bm{W}_k^j\bm{\Check{\mathcal{F}}}_{t-\Delta t}^j/\sqrt{C} + \bm{0} \right)
	\end{aligned}
    \end{equation}
where $\bm{W}_q$, and $\bm{W}_k$ are of learnable weights. $\bm{\Gamma}(i, j)_t^{t-\Delta t}$ is affinity score between $i^{th}$ object in frame $\bm{I_t}$ and $j^{th}$ object in frame $\bm{I_{t- \Delta t}}$. Similar to spatial information flow module, temporal positional encoding is set as $\bm{0}$. Considering that the objects in the image pair may not have a corresponding relationship, we learn a extra row and column for un-identified targets, following  the DAN~\cite{DBLP:journals/pami/SunASMS21}. We simple add a FFN to weights of cross-attention and estimate affinity matrix $\bm{\Gamma}_t^{t-\Delta t} \in \mathbb{R}^{K+1 \times K+1}$
, which will be trained as a unimodal matrix for one-to-one matching.

The temporal information flow branch can generate tracking information and guide the temporal aggregation of appearance and geometric information. The temporal aggregation module then models the temporal transition of targets to predict box smoothness refinement and the time-related variable(\emph{e.g., velocity, motion attribute}). Thus, the mechanism of temporal information flow can be summarized as:
    \begin{equation}
	\label{quta}
	\begin{aligned}
	\bm{\tilde{\mathcal{F}}}_{t}^i = MultiHeadAttn(\bm{\Hat{\mathcal{F}}}_{t}^i, \bm{\Check{\mathcal{F}}}_{t-\Delta t}, \bm{\Check{\mathcal{F}}}_{t-\Delta t}, \bm{0})
	\end{aligned}
    \end{equation}
where $\bm{\tilde{\mathcal{F}}}_{t}^i \in \mathbb{R}^{K \times d}$ is the aggregation features of $i^{th}$ target at time $t$. The final prediction is computed by a FFN which consist of 3-layer MLP. The FFN predicts the velocity $\bm{V}_t = \{\bm{V}_t^i \in \mathbb{R}^3 |i=1,\dots,K\}$, motion attribute $\bm{M}_t = \{\bm{M}_t^i \in \mathbb{R}^3 |i=1,\dots,K\}$, and box smoothness refinement value $\Delta \bm{\mathcal{B}} = \{\Delta \bm{\mathcal{B}}^i_t = \Delta X_t^i, \Delta Y_t^i, \Delta Z_t^i, \Delta W_t^i, \Delta H_t^i, \Delta L_t^i | i=1,\dots, K \}$. 

\subsection{Training Loss}
We divide the multi-task loss into three parts: monocular object 3D detection loss $\mathcal{L}_{Mono3D}$, the tracking loss $\mathcal{L}_{\mathcal{T}}$, and temporal-consistency loss $\mathcal{L}_{Cons}$. 

\textbf{1). Monocular object 3D detection loss.} We adopt the same loss as KM3D~\cite{li2020monocular}:
\begin{equation}
	\label{eq:a}
	\begin{aligned}
	\mathcal{L}_{Mono3D} = \mathcal{L}_{m} + \mathcal{L}_{kc} + \mathcal{L}_{D} + \mathcal{L}_{O} + \mathcal{L}_{T}
	\end{aligned}
\end{equation}
where the $\mathcal{L}_{m}, \mathcal{L}_{kc}, \mathcal{L}_{D}, \mathcal{L}_{O}, \mathcal{L}_{T}$ are main center loss, keypoint loss, dimension loss, orientation loss and position loss respecitively. 

\textbf{2). Tracking loss.} 
Benefit to our explicit modeling of the target appearance features and geometric features in spatial-temporal space, we can design a simpler and more effective loss function then DAN~\cite{DBLP:journals/pami/SunASMS21} to constrain the affinity matrix to generate a unimodal response. Specifically, we first perform row- and column-wise \emph{softmax} to affinity matrix $\bm{\Gamma}_t^{t-\Delta t} \in \mathbb{R}^{K+1 \times K+1}$, and results association matrix $\bm{\Hat{\Gamma}}_{t, t-\Delta t}$ and $\bm{\Hat{\Gamma}}_{t-\Delta t, t}$. The $i^{th}$ row of the matrix $\bm{\Hat{\Gamma}}_{t, t-\Delta t}$ associates the $i^{th}$ objects in frame $\bm{I}_{t}$ to $K+1$ identities in frame $\bm{I}_{t-\Delta t}$, where $+1$ presents the un-identified targets in frame $\bm{I}_{t-\Delta t}$. The matrix $\bm{\Hat{\Gamma}}_{t-\Delta t, t}$ signify similar associations from $\bm{I}_{t-\Delta t}$ to $\bm{I}_{t}$.
We then multiply these two matrices $\bm{\tilde{\Gamma}}_t^{t- \Delta t} =  \bm{\Gamma}_{t, t - \Delta t} \times \bm{\Hat{\Gamma}}_{t - \Delta t, t}$ and use cross-entropy losses for training:
\begin{equation}
	\label{eq:a}
	\begin{aligned}
	\mathcal{L}_{tracking} = - \sum_{j=1}^{K} \sum_{i=1}^{K+1}  *\bm{\Gamma}_{t}^{t-\Delta t}(i, j)\log(\bm{\tilde{\Gamma}}_{t}^{t-\Delta t}(i, j)) \\
	\end{aligned}
\end{equation}
where $^*\bm{\Gamma}_t^{t-\Delta t}$ presents the ground truth association matrix, and $^*\bm{\Gamma}_t^{t-\Delta t}(i, j)=1$ indicates that the $i^{th}$ identities in the frame $\bm{I}_{th}$ and the $j^{th}$ object in the frame $\bm{I}_{t}$ are the same object. Fig.~\ref{fig:trackingscore} show a sample frames from nuScenes video to illustrate the processing of tracking loss. 

\textbf{3). Temporal-consistency loss.} The design of most functions hopes to check the stability in temporal to ensure a smooth transition. However, previous approaches calculated detection results independently for each frame and failed to constrain a trajectory's inner consistency in the network. We propose an auxiliary loss to clarify temporal topology between frames during the training phase to ensure trajectory smoothness. Specifically, we first add the smoothness refinement box to the outputs of our monocular 3D detector to generate final box parameters and then calculate the relative 3D corner distance of the same object in a different frame. The ground truth then supervises these corner distances: 
    \begin{equation}
	\label{eq:b}
	\begin{aligned}
	 &\mathcal{L}_{Cons} =  \sum_{\zeta=1}^{\tau} \sum_{i=0}^{n_t} \sum_{j=0}^{n_{t-k}}
	 \left| \left| \mathcal{C}_3(\bm{B}_{t}^i + \Delta \bm{B}_{t}^i) \right. \right.\\
	 &\left. \left. - \mathcal{C}_3(\bm{B}_{t-\zeta}^j + \Delta \bm{B}_{t-\zeta}^j) \right|_2 - \left| \mathcal{C}_3(B^*_t)-\mathcal{C}_3(B^*_{t-\zeta})\right|_2 \right|_2
	\end{aligned}
    \end{equation}
where $\mathcal{C}_3$ map the parameters of the 3D box to 8 corners. $B^*$ presents ground truth. Interestingly, although every potential target is supervised by corresponding ground truth, the temporal-consistency loss can still minimize the relative location. Fig.~\ref{fig:cons} shows a sample to explain that. 

Finally, the Time3D is supervised in an end-to-end fashion with the multi-task loss combination:
    \begin{equation}
	\label{eq:b}
	\begin{aligned}
	 \mathcal{L} = \mathcal{L}_{Mono3D} + \mathcal{L}_{Tracking} + \mathcal{L}_{Cons}
	\end{aligned}
    \end{equation}
\begin{figure}[htb]
\setlength{\abovecaptionskip}{0pt}
	\begin{center}
		\includegraphics[width=0.9\linewidth]{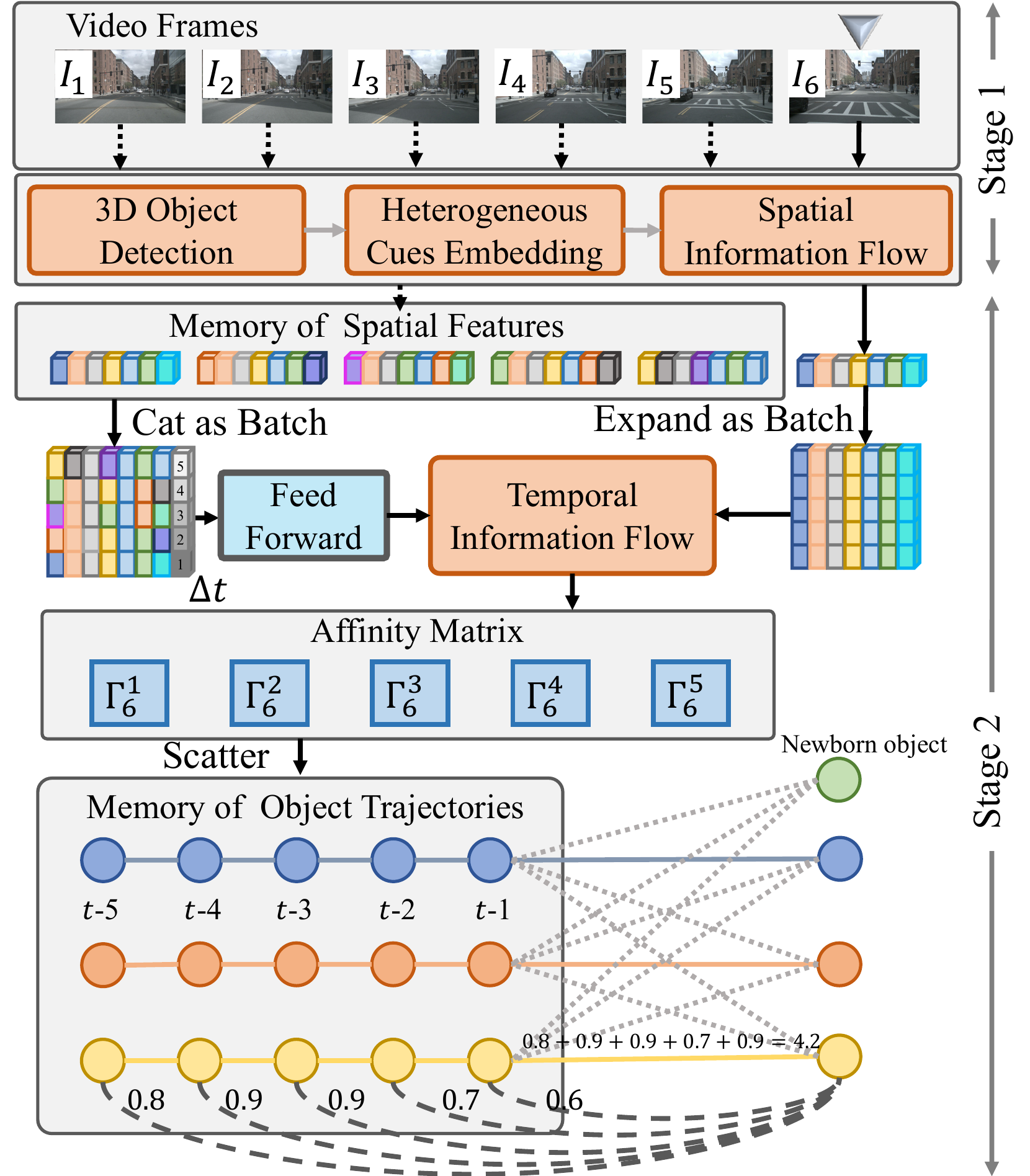}
	\end{center}
	\caption{\textbf{Inference of tracking.} Each frame of images is first fed into 3D object detection, embedding extractor, and spatial information flow module to generate spatial features. We then equip an explicit memory to store the past trajectories with corresponding spatial features. Given the current spatial features, only the fast and straightforward temporal information flow is used to compute the similarity score.}
	\label{fig:tracking}
	\vspace{-0.4cm}
\end{figure}
\subsection{Inference of Tracking}\label{sec:it}
The inference of tracking is shown in Fig.~\ref{fig:tracking}.
We first perform the 3D object detection, heterogeneous cues extractor, and spatial information flow sequentially at each timestamp. Next, the spatial features are stored with their time stamps. Given the current frame image $I_t$ with its spatial features, the affinity matrix is computed by forward passing through temporal information flow. In order to reduce the ID switch, we follow the DAN~\cite{DBLP:journals/pami/SunASMS21} to calculate the affinity matrix of the objects in the current frame and the spatial features stored in all the trajectories and sum them as the similarity score between the objects and the trajectories. Finally, the Hungarian algorithm~\cite{1962Algorithms} is adopted to get the optimal assignments. When running the assignment process frame-by-frame, object trajectories are yielded. Thus each frame image is passed through a high-weight network of 3D detection, embedding extractor, and spatial information only once, but
the stored spatial features are used multiple times through light-weight temporal information flow to compute similarity score. Time3D, therefore, can run in real-time.
\section{Experiments}
\subsection{Dataset}
We evaluate Time3D on a large-scale popular nuScenes~\cite{Wang_2021_ICCV} dataset for autonomous driving. nuScenes is composed of 1000 scenes video, with 700/150/150 scenes for \emph{train}, \emph{validation}, and \emph{testing} set, respectively. nuScenes collected data from Boston and Singapore, in both day and night, and different weather conditions. Each video contains images from six cameras forming an entire $360^{\circ}$ field of view in a 2HZ keyframe sample. Finally, 1.4M 3D bounding box annotations from 10 categories are contained. Considering that nuScenes includes sequence monocular images, 3D annotations, tracking ID, and the pose of each frame simultaneously, we use it as the benchmark to validate the performance of our method. 
\begin{table*}[!ht]
    \scriptsize
	\begin{center}
			\caption{\textbf{3D Tracking and 3D Object Detectiom Performance nuScenes \emph{test} set.} $\dagger$ represent it use the AB3DMOT~\cite{DBLP:journals/corr/abs-1907-03961} as the tracker. Time3D$\ddagger$ trained in no end-to-end manner.}
		    \begin{tabular}{ c | c | c | c | c | c | c | c |c | c | c | c | c | c }
			\hline
			\multirow{3}{*}{Method} & \multirow{3}{*}{Modality} &\multirow{3}{*}{Time} & \multicolumn{4}{c|}{3D Multi-Object Tracking} & \multicolumn{7}{c}{3D Object Detection}\\
			\cline{4-14}
			 & & & AMOTA  & AMOTP   & MOTA  & MOTP  & mAP & mATE & mASE & mAOE & mAVE & mAAE & NDS  \\
			&&&(\%) $\uparrow$&(m) $\downarrow$&(\%) $\uparrow$&(m) $\downarrow$&(\%)$\uparrow$&(m)$\downarrow$&(1-iou)$\downarrow$&(rad)$\downarrow$&(m/s)$\downarrow$&(1-acc)$\downarrow$&(\%)$\uparrow$\\
			\hline
			Megvii$\dagger$~\cite{DBLP:conf/cvpr/CaesarBLVLXKPBB20} & LiDAR & - & 15.1 & 1.50  & 15.4 &\textbf{0.40}& \textbf{52.8} & \textbf{0.30} & \textbf{0.25} & \textbf{0.38} & \textbf{0.25} & 14.0 & \textbf{63.3} \\
			PointPillar$\dagger$~\cite{DBLP:conf/cvpr/CaesarBLVLXKPBB20} & LiDAR & - & 2.9 & 1.70 & 4.5 & 0.82 & 30.5 & 0.52 & 0.29 & 0.50 & 0.32 & 37.0 & 45.3 \\
			\hline 
			
			CenterNet~\cite{zhou2019objects} & Camera & -  &  - & - & - & - & 33.8 & 0.66 & 0.26 & 0.63 & 1.63 & 14.2 & 40 \\
		    FCOS3D~\cite{Wang_2021_ICCV} &Camera& - &  - & - & - & - & 35.8 & 0.69 & 0.25 & 0.45 & 1.43 & 12.4 & 42.8 \\
			\hline
			CenterTrack~\cite{DBLP:conf/eccv/ZhouKK20}& Camera & 45ms & 4.6 & 1.54  & 4.3 & 0.75 & - & - & - & - & - & - & - \\
			DEFT~\cite{DBLP:journals/corr/abs-2102-02267} & Camera & - & 17.7 & 1.56  & 15.6 & 0.77 & - & - & - & - & - & - & - \\
			\hline
			MonoDIS$\dagger$~\cite{DBLP:conf/cvpr/CaesarBLVLXKPBB20} &Camera& - &1.8 & 1.79  & 2.0 & 0.90 & 30.4 & 0.738 & 0.263 & 0.546 & 1.553 &  13.4 & 38.4 \\
			Time3D$\ddagger$ & Camera & {50ms} & {15.6} & {1.49} & {14.1} & 0.78 & {32.9} & {0.716} & {0.250} & {0.511} & {1.647} & {14.8}
			& {39.9} \\
			Time3D & Camera & \textbf{38ms} & \textbf{\textcolor{black}{21.4}} & \textbf{\textcolor{black}{1.36}} & \textbf{\textcolor{black}{17.3}} & 0.75 & \textcolor{black}{31.2} & \textcolor{black}{0.732} & \textcolor{black}{0.254} & \textcolor{black}{0.504} & \textcolor{black}{1.523} & \textbf{\textcolor{black}{12.1}}
			& \textcolor{black}{39.4} \\
			\hline
		   \end{tabular}
	       \label{tab:tracking}
	\end{center}
	\vspace{-0.4cm}
    \end{table*}
\subsection{Implementation Details}
In mono3D, we follow the KM3D~\cite{li2020monocular} to report the results with the DLA-34 backbone. We stack three self-attention layers in spatial information flow, and stack four cross-attention layers in temporal information flow, in which we compute affinity matrix in $2^{th}$ cross-attention without \emph{softmax}.
Time3D is trained without the KM3D pre-trained model but with ImageNet pre-trained model as initialization. We use the augmentation of shifting and scaling. We resize the image from $ 900 \times 1600 $ to $448 \times 800$ for fast training. 
The batch size is 80 in 8 2080Ti GPUs,
with ten images per GPU. We train Time3D for $200$ epochs with a starting learning rate of $1.25e-4$ and reduce it by $ \times 10 $ at epoch 90 and 120.
We only associate the past five frames to the current frame for the fast running speed.
\subsection{Comparison with state-of-the-art}
We report four official evaluation metrics~\cite{DBLP:conf/avss/KutschbachBES17} for tracking quantitative analyses: Average Multi-Object Tracking Accuracy (AMOTA), Average Multi-Object Tracking Precision (AMOTP), Multi-Object Tracking Accuracy (MOTA), and Multi-Object Tracking Precision (MOTP), while reporting the seven official evaluation metrics for 3D detection: Average Precision (AP), Average Translation Error (ATE), Average Scale Error (ASE), Average Orientation Error (AOE), Average Velocity Error (AVE), Average Attribute Error (AAE), and nuScenes detection score (NDS). We only compare the online (no peeking into the future) tracking methods from published approaches for a fair comparison. As shown in Tab. \ref{tab:tracking}, Time3D outperforms all the other online trackers with different metrics while maintaining outstanding running speed. As we all know, the MOT performance is highly relay on the accuracy of detection. We therefore list the results of two popular LiDAR-based 3D detector~\cite{DBLP:journals/corr/abs-1908-09492,DBLP:conf/cvpr/LangVCZYB19} with baseline tracker AB3DMOT~\cite{DBLP:journals/corr/abs-1907-03961}. Megvii achieves high detection accuracy by employing LiDAR point cloud, but Time3D outperforms the Megvii by a large margin for 3D tracking, in which AMOTA improves 41.7\%, AMOTP of 9\%, MOTA 11\%. We also perform a Time3D with no end-to-end training, in which we separately train KM3D, Re-ID extractor, and spatial-temporal module, detailed in supplementary. End-to-end Time3D improves 37\% for AMOTA and  22\% for MOTA. 
All these shows that the power of Time3D, which jointly learns detection and tracking in an end-to-end manner, can outperform those methods that use detection and tracking as two different steps, even if these methods have more powerful detection accuracy. Tab. \ref{tab:tracking} show Megvii has a smaller MOTP to Time3D, which is because LiDAR-based Megvii has a higher accuracy to a translation error.

Time3D is proposed to combine 3D detection and 3D tracking in a unified framework and make autonomous driving more practical. We, therefore, employ a lightweight monocular 3D detection method, KM3D, for the trade-off between speed and accuracy. The 3D detection accuracy of Time3D still has a gap to LiDAR-based methods Megvii and has the competitiveness to single task methods CenterNet~\cite{zhou2019objects} and FCOS3D~\cite{Wang_2021_ICCV}. The use of a more powerful 3D detector can improve the performance of 3D detection, and we will take it as future work. Note that Time3D$\ddagger$ achieves higher accuracy in the 3D detection benchmark, and we will analyze this phenomenon in the ablation study.


\subsection{Qualitative Results}
\begin{figure*}[htb]
\setlength{\abovecaptionskip}{0pt}
	\begin{center}
		\includegraphics[width=0.8\linewidth]{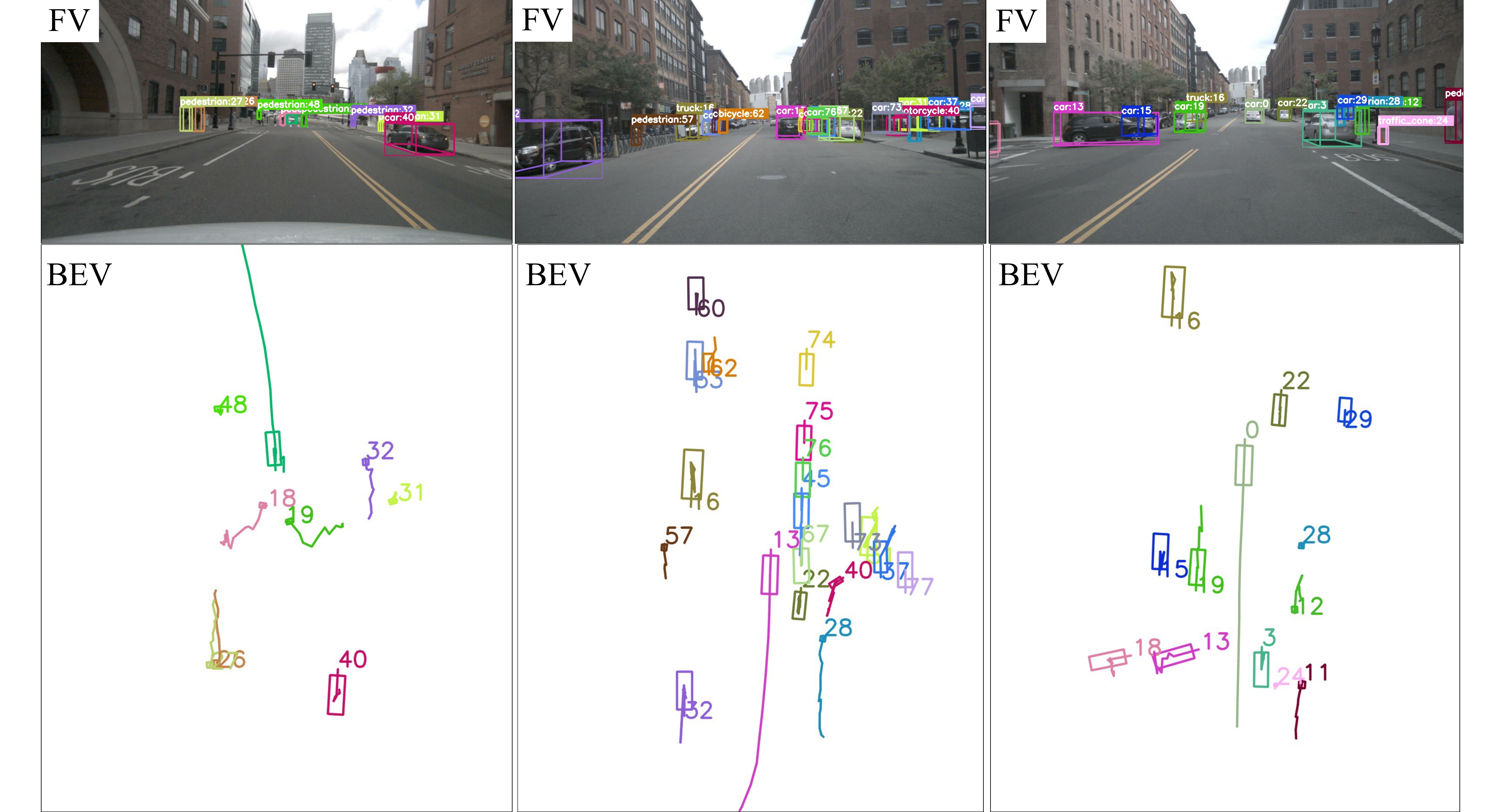}
	\end{center}
	\caption{\textbf{Qualitative Results.} We visualize the tracking ID, 2D box, 3D box, category in front view in $1^{th}$ row. The $2^{th}$ row of Bird's eye view (BEV) show the 3D box with the center point trajectory of the past 15 frames. Different colors represent objects with different tracking IDs.}
	\label{fig:vis}
	\vspace{-0.4cm}
\end{figure*}
Fig.~\ref{fig:vis} shows some qualitative results of Time3D.  
Time3D can predict exciting tracking effects and output smooth trajectories, even for occluded or high-speed moving targets. It is worth noting that Time3D only uses monocular images as input and outputs 3D trajectories. Small targets such as pedestrians have relatively rough trajectories related to the 3D detector. It can be replaced with a more robust 3D detector to generate a more accurate 3D position. More qualitative results can be seen in the supplementary.
\subsection{Ablation Study}
For a fair comparison, all experiments are trained in \emph{training} set and tested in \emph{val} set following the nuScenes official setting. 
\begin{table}[!ht]
    \setlength{\abovecaptionskip}{8pt}
	\scriptsize
	\begin{center}
			\caption{\textbf{Ablation for different cues} Performance on nucenes $val$. $\star$ represent directly encoding to 2D or 3D box.}
		\begin{tabular}{ c  | c | c | c |c|c|c|c}
			\hline
			Cls & 2D & 3D &  Re-ID&  AMOTA  & AMOTP   & MOTA  & MOTP  \\
			\hline
			\checkmark&  & & & 2.4 & 1.79 &2.7 &0.74\\
			          & \checkmark& & &8.9 & 1.68& 10.1 &0.76\\
			& & \checkmark& & 11.2 & 1.61 & 12.8 & 0.76\\
		    & & & \checkmark& 16.5 & 1.54 & 15.3 &0.82\\
		    & &\checkmark& \checkmark& 24.7 & 1.41 & 19.6&0.82\\
		    &\checkmark& \checkmark& \checkmark& 25.8 &1.39 & 20.4  &0.83\\
		    \checkmark& \checkmark&\checkmark & \checkmark&26.0 & 1.38 & 20.7 &0.82\\
		    \hline
		    \checkmark& \checkmark $\star$&\checkmark $\star$ & \checkmark& 19.0 & 1.48 & 17.1 &0.82\\
			\hline
		\end{tabular}
		\label{tab:hce}
	\end{center}
	\vspace{-0.4cm}
    \end{table}

\textbf{Heterogeneous Cues Embedding.}
We first examine the effects of different cues, such as Re-ID features, the embedding of 2D box, 3D box, and category. As shown in Tab~\ref{tab:hce},
for the single cue testing, using only Re-ID features achieves the highest 16.5 AMOTA. Adding the other cues can further boost the performance. 
The best performance is achieved by combining all heterogeneous cues. We observe that the biggest improvement occurs in adding the 3D Box cue, which demonstrates that the 3D location is important to 3D object tracking. We expect to inform future research to focus on 3D information to improve the tracking performance further.

The geometric embedding extractor learns a compatible feature representation for 2D boxes and 3D boxes. We investigate its effect by comparing the solutions that directly encode 4 Dof and 7 Dof parameters of the 2D Box and 3D Box with the same MLP layer as the proposed geometric embedding extractor. As shown in the last row of Tab \ref{tab:hce}, direct encoding has dropped performance significantly, which demonstrates that an explicit geometry encoding method is more conducive to the learning of the network.

\begin{table}[!ht]
    \setlength{\abovecaptionskip}{8pt}
	\scriptsize
	\begin{center}
			\caption{\textbf{Ablation for with or without Re-ID feature.} Performance on nucenes $val$.}
		\begin{tabular}{ c  | c | c | c |c|c}
			\hline
			Setting &  mAP&  mATE  & mASE   & mAOE  & NDS  \\
			\hline
			w/& 29.1  &  0.79&  0.24& 0.48 & 39.0\\
			w/o& 32.2 & 0.75 & 0.26 & 0.49 & 39.5\\
			\hline
		\end{tabular}
		\label{tab:reidcons}
	\end{center}
	\vspace{-0.4cm}
    \end{table}
In addition, we observe that MOTP has a little worse when introducing the Re-ID feature. We, therefore, performed the detection experiment and found that introducing the Re-ID harms the detection performance. The results are shown in Tab.~\ref{tab:reidcons}. This also explains that the detection accuracy of Time3D$\ddagger$ is better than Time3D. The degeneration may be caused by the contradiction between the invariance of Re-ID "identity" and the variance of detection.  

\begin{figure}[htb]
\setlength{\abovecaptionskip}{0pt}
	\begin{center}
		\includegraphics[width=1\linewidth]{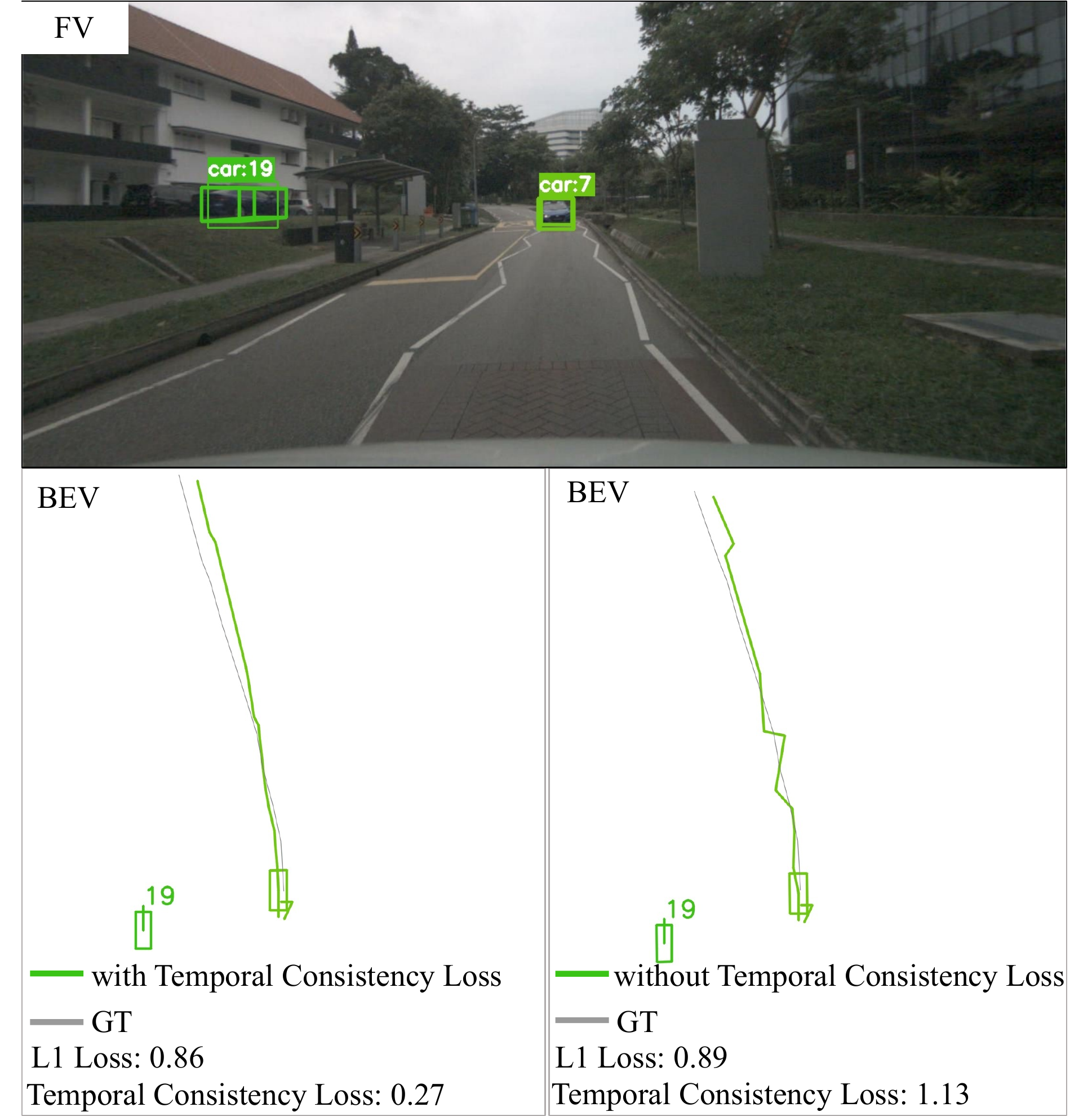}
	\end{center}
	\caption{\textbf{A sample to illustrate the temporal-consistency loss.} We show two results predicted by Time3D with and without temporal-consistency loss. They have the same corner $L_1$ loss but have different temporal-consistency loss. Time3D with temporal-consistency loss generates a smoother trajectory. }
	\label{fig:cons}
	\vspace{-0.4cm}
\end{figure}
\begin{table}[!ht]
    \setlength{\abovecaptionskip}{8pt}
	\scriptsize
	\begin{center}
			\caption{\textbf{Ablation for Spatial-Temporal Information Flow}}
		\begin{tabular}{ c  | c | c | c | c }
			\hline
			Setting &  AMOTA  & AMOTP   & MOTA  & MOTP \\
			\hline
			w/o Spatial-Temporal & 17.9 & 1.43 & 19.0 & 0.81\\
			w/ Spatial & 23.9 & 1.27 & 14.7 & 0.82 \\
			w/ Saptial-Temporal & 26.0 & 1.38 & 20.7 & 0.82\\
			\hline
		\end{tabular}
		\label{tab:STIF}
	\end{center}
	\vspace{-0.4cm}
    \end{table}
    
\textbf{Spatial-Temporal Information Flow.}

Tab.~\ref{tab:STIF} examines the effects of the spatial-temporal information flow module in improving the tracking accuracy. To avoid the influence caused by additional network parameters, we replace the spatial and temporal information module with 6 MLP layers and cosine similarity. The proposed spatial-temporal information flow module improves by 45\% in terms of AMOTA. At the same time, it can also predict velocity, motion attribute, and box smoothness refinement, indicating the \emph{spatial-temporal information flow} can more explicitly detect the spatial-temporal topological relationship between objects.


\textbf{Temporal-Consistency Loss.}
We show a sample in Fig.~\ref{fig:cons} to illustrate the effect of temporal-consistency loss. The trajectory with temporal consistency-loss is smoother than the trajectory temporal-consistency loss. They have a similar $L_1$ mean distance loss to the ground truth but have different temporal-consistency loss. We propose the temporal-consistency matrices to evaluate the smoothness of trajectory. The details can be seen in supplementary. 

\section{Conclusion}
This work proposes a novel framework to jointly learn 3D object detection and 3D multi-object tracking from only monocular video while running in real-time. Our framework encodes the heterogeneous cues, including category, 2D Box, 3D Box, and Re-ID feature, to compatible embedding. The transformer-based architecture performs spatial-temporal information flow to estimation trajectory, optimized by temporal-consistency loss to smoother. On the nuScenes dataset, the proposed Time3D achieves state-of-the-art tracking performance while running in real-time. The Time3D may inspire future researchers to combine 3D tracking and 3D detection in a unified framework, and to encode more 3D information to make vision-based autonomous driving more practical.
{\small
\bibliographystyle{ieee_fullname}
\bibliography{ref}
}

\end{document}